# Go Big or Go Home: Simulating Mobbing Behavior with Braitenbergian Robots

Ellie Sanoubari
*Artificial Life: Embodied Intelligence*
*Fall 2019*
esanouba@uwaterloo.ca

*Abstract*— We used the Webots robotics simulation platform to simulate a dyadic avoiding and mobbing predator behavior in a group of Braitenbergian robots. Mobbing is an antipredator adaptation used by some animals in which the individuals cooperatively attack or harass a predator to protect themselves. One way of coordinating a mobbing attack is using mobbing calls to summon other individuals of the mobbing species. We imitated this mechanism and simulated Braitenbergian robots that use mobbing calls when they face a light source (representing an inanimate predator) and mob it if they can summon allies, otherwise, they escape from it. We explore the effects of range of mobbing call (infinite range, mid-range and low-range) and the size of the robot group (ten robots vs three) on the overall success of mobbing. Our results suggest that both variables have significant impacts. This work has implications for simulations of action selection in artificial life and designing control architectures for autonomous agents.

## I. INTRODUCTION

Anti-predatory adaptations are defense mechanisms that living things have developed through evolution to help them in their struggles against their predators. For example, *thanatosis* [1] or playing dead is one such mechanism that helps a prey ward off a predator's attack. In this work, we take inspirations from an antipredator adaptation called *mobbing* (See Fig 1).

Mobbing is when a group of prey species join forces and cooperatively attack a predator or harass it (to cause it to move out of the vicinity, [2]). Some species (e.g. many birds) use mobbing calls for signalling each other to carry out the attack. Mobbing calls are usually in a frequency range that is heard by other members of the mobbing species but is not heard by the predator to help them catch the predator off guard [2]. In this work, we drew inspirations from mobbing mechanism in animals and simulated robots that mob an inanimate predator.

The behavior design of these robots was inspired by the fictional vehicles described by Braitenberg [3]. We built to his thought experiments to design behaviors for the robots in this simulation that may be interpreted as "fear", "knowledge", "aggression" and "logic".

The robots in this experiment have two primary behaviors: i) escaping the predator when there are no other robots around which can be interpreted as fear, and ii) mobbing the predator when there are other robots to help which can be interpreted as "knowledge" of other robots being around and "aggression" towards the predator. The robots can switch their behavior from avoiding the predator to mobbing it when the situation is appropriate, which may infer their "logic".

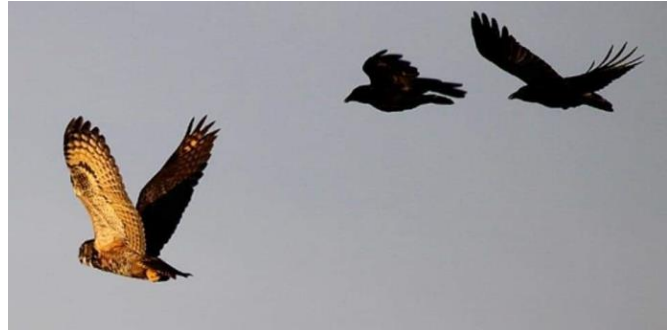

Fig. 1. Crows mobbing an owl, source: [25]

We used Cyberbotics Webots [4] platform to simulate this work tested the impacts of the range of robots' mobbing calls and group size of robots on their overall mobbing success. We explored mobbing range in three levels: no range, medium range, and small range with groups of ten and three robots in a 2x3 experimental design. We conducted 60 simulations over 10 randomized simulation worlds and found significant main effects of both factors and some interaction effects.

In the following sections, we outline related work in behavior design in artificial life, Braitenberg simulations and extended architectures, bio-inspired group behaviors in artificial life, and simulating cooperative anti-predatory behaviors. We then introduce our research questions and hypothesis and detail our methodology, followed by our experiment design and analysis approach. Finally, we present our quantitative findings and conclude with a critical evaluation of our work.

## II. LITERATURE REVIEW

Langton defines artificial life as life-like behavior made by man rather than by nature, and argues that study of artificial life complements the empirical research in biology and contributes to it by broadening the horizons of research from "life-as-we-know-it", to "life-as-it-could-be"[5]. Since he coined the term in the late 80s, he inspired many researchers to study artificial life.

He puts forward the argument that complex global behavior is synthesized from non-linear interactions happening at the local level [5]. In the same line of thought, Braitenberg describes a set of fictional vehicles to demonstrate how simple mechanical and electronic components can be synthesized to form complex agents and imitate intelligent behaviors such as fear or aggression [3]. Similarly, related work offering design principles for autonomous agents states that intelligence is emergent from "parallel, loosely coupled processes" that run asynchronously with no or little centralized resources [6]. We build on this body of work and combine processes on simple electronic components such as sensors and transmitters to form relatively complex behaviors.

Braitenberg's thought experiments have inspired a lot of researchers. Prior work has used various virtual [7], or physical tools [8] (e.g. electronic leggo bricks) to carry out simulations of simple Braitenberg vehicles. One project has developed a Braitenbergian controller for a fish robot that uses pressure sensors to navigate itself in the flow [9].

Further, related work has attempted to extend Braitenberg architectures [10], [11]. For example, Lambrinos and Scheier extended Braitenberg's architecture to include action selection capability to enable their robots to make a decision when there are more than one possible tasks [12]. Similarly, the robots in this simulation are designed to make a decision to switch their behavior from avoiding the predator to mobbing it when they see fit.

Other prior work that has attempted to extend and formalize Braitenbergian models of taxis highlights that it is important to consider time-varying stimulus and dynamic sensor response [13]. An example of such stimuli can be found it the work of Lilienthal and Duckett-- they have developed smelling Braitenbergian robots with gas sensors to study localization of static odor source in unstructured environments with turbulent distribution of odur molecules [14]. Even though we do not directly employ time-varying stimuli in our simulation, we impose an emitter range on moving robots which could be interpreted as time-varying stimuli from a receiver's point of view.

With increasing interest in bioinspiration in robotics, systems of multiple autonomous mobile agents that exhibit group behavior have received special attention [15]–[17]. For example, one project has developed and studied a group of robots that collaborate to gather random objects and compared them to ant colonies [18].

Furthermore, some research groups have implemented experimentation tools for supporting this line of research. Werner and Dyer have built a simulation platform (BioLand) for studying evolution of herding behavior in artificial animals [19]. Related work has used one such tool to simulate artificial fish schooling and study their anti-predatory mechanisms [20]. Our project follows the same line of work by simulating another well-known anti-predatory behavior.

Finally, some prior work has experimented with cooperative anti-predatory behaviors similar to mobbing. Davis and Arkin used models of risk and dishonesty from games theory to simulate mobbing by autonomous agents [21]. Another work concerned with mobile robot troops has presented robotic system that exhibits a cooperative behavior for defense against security invasion in areas under surveillance [22]. Building on this body of work, we designed and simulated a non-linear anti-predatory adaptation that includes both escaping and mobbing the predator.

### III. RESEARCH QUESTIONS AND HYPOTHESES

This work addresses three primary research questions about the possibility of simulating a mobbing scenario with Braitenbergian robots (Q1), and the effects of the range of mobbing call (Q2) and group size of robots (Q3) on mobbing success.

*Q1. Is it possible to simulate avoiding/mobbing predator behaviors using Braitenbergian robots?*

Asking questions about the possibility of simulating behaviors of living things is central to artificial life research. It is also important to consider how realistic the simulated scenarios are. That is, it should be possible to adjust the difficulty of simulated scenarios such that, just like mobbing species in real life, robots face different levels of difficulties do not always succeed in mobbing. We hypothesize that:

[H1a] it is possible to simulate a dyadic avoiding/mobbing antipredator behavior with Braitenbergian robots.

[H1b] it is possible to manipulate some simulation factors to make a range of mobbing scenarios with different levels of difficulty.

*Q2. How does the range of robots' mobbing call affect the success of simulated mobbing?*

When animals attempt to coordinate a mobbing attack, their mobbing call can only be heard by other individuals in a certain distance. To make the mobbing simulation more realistic, it is important to impose a range to robot's mobbing call. We hypothesize that:

[H2a] compared to having unlimited range, imposing a call range will affect overall mobbing success negatively.

[H2b] robots will higher mobbing range will succeed in coordinating mobbing more often.

[H2c] imposing a mobbing range will have a bigger impact in terms of mobbing success on smaller robot groups compared to bigger groups.

*Q3. How does the group size of robots impact the success of simulated mobbing?*

Mobbing is a group-work, and it is important to consider how its qualities will change with increasing or decreasing the group size. Intuitively, we know that sometimes a group can be too small or too big for completing a certain task. In the context of this simulation, we hypothesize that:

[H3a] bigger groups of robots will be successful in coordinating a mobbing more often.

[H3b] robots in bigger groups will have less participation rates in mobbing.

In our hypotheses we define overall mobbing success to include both the overall number of unanimous or partial successful mobbing attempts, and the percentage of participation of robots in their group's mobbing attempt (the higher, the more successful).

### IV. METHODOLOGY

We designed and simulated Braitenbergian robots that would sense and avoid a source of light (representing an inanimate predator) while trying to coordinate a mobbing attempt. If they could coordinate a mobbing attempt, they would change their behavior and mob the predator instead of avoiding it. We explored the effects of mobbing range and robot group size in a 2x3 experiment. Each experimental condition was tested on ten different simulation worlds. The initial locations of all the objects were randomized. We detail our methodology below.

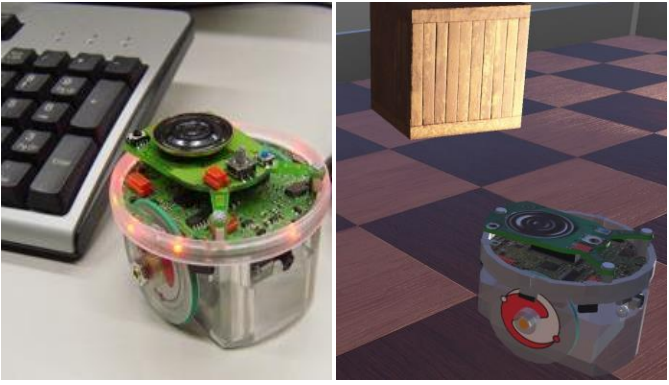

Fig. 2. Left: e-puck robot, source: [4], Right: simulated e-puck robot

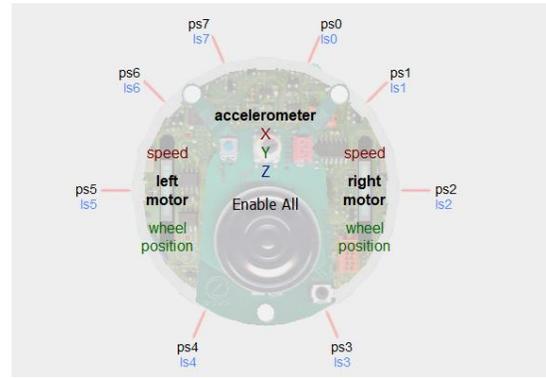

Fig. 3. E-puck robot sensors, screenshot from Webots Software [4]

*A. Implementation*

This simulation was implemented in Webots robotic simulations software [4]. Webots is an open-source 3D simulation platform that can be used for modeling and programming robots. Using this platform, we designed a 3D environemnt (aka worlds) containing several objects including obstacles, a light source, and Braitenbergian robots (See code in appendix A, and simulation files in the email attachments).

*1) Objects*

We used a model of GCTronic's e-puck [23] in this simulation (See Fig 2). E-puck is an open-source miniature two-wheeled robot (7.4 cm in diameter). In this simulation, we used e-puck's ambient sensors to sense the predator (represented by a light source), and distance sensors to avoid the obstacles. We also used this robot's emitter and receiver to send and receive mobbing calls. We used both of e-puck's wheel motors to move the robot around and make left or right turns. No other components or sensors of this robot were used in the simulations.

We simulated the inanimate predator using a point-light node. This node simulates one point as light source that emits light equally in all directions. This light was placed directly on the simulated floor (i.e., at height zero). We also increased the intensity of this light to double the default value for better visibility. We argue that the design of the predator as an inanimate object makes sense because when animals make a mobbing call it usually has the evolutionary advantage that it cannot be detected by their predator. This gives the mobbing species the opportunity to catch their predator by surprise.

We used wooden boxes to simulate obstacles in the simulation environment. The boxes are 10x10x10 centimeters and opaque (i.e. not transparent to light). The boxes were assigned a mass value so that the robots would not be able to push them around (See Fig 2).

*2) Braitenbergian Behaviors*

We programmed the robot controllers to use sensory input for making decisions about their directions. We used the sum of sensory inputs on the right and left side of the robots to detect light/distance on each side (see Fig 3). That is, the sensors were treated as 2 light sensors and 2 distance sensors. The platform enabled us to assign different velocities to each wheel motor, which was used for making right and left turns.

We implemented three different Braitenbergian behaviors: avoiding predator, mobbing predator, and avoiding obstacles. The mobbing and avoiding predator behaviors were mutually exclusive and would not happen at the same time. However, the avoiding obstacles behavior was consistent through the simulation. The avoiding obstacles behavior was designed to be easily overpowered by the other two behaviors so it would be reconcilable with them. For example, if two robots that were trying to mob a predator hit one another, their obstacle avoidance would not interfere with their mobbing. This was done by programming the robots to make more aggressive decisions in their antipredator behaviors than their obstacle avoidance (e.g. by making sharper and faster left or right turns).

*a) Avoiding obstacles:* we programmed the robots to always avoid obstacles. That is, if a robot faced an obstacle, it would turn around and continue moving in another direction. This was implemented so that the robots would not stop if they hit a wall or a box and simulation would go on. We programmed the controllers such that if the robot detected an obstacle on one side, it would assign a slightly lower speed to the opposite wheel causing the robot to slowly turn to the opposite direction of the obstacle.

*b) Avoiding predator:* when the robots sense a certain intensity of light, this behavior makes them turn them away from the light and continue moving in an opposite direction. This Braitenbergian behavior can be interpreted as "fear" [3] and simulates how an individual from a mobbing species would normally behave in front of a predator. We programmed the controllers such that if the source of light was detected on one side, the controller would assign a lower speed to the opposite wheel causing the robot to quickly turn to the opposite direction of the light.

*c) Mobbing predator:* when this behavior is initialized, the robots stop their avoiding predator behavior and instead turn towards the light and continue moving in its direction. When they hit the light source and pass it, they turn around and move in its direction again. This Braitenbergian behavior simulates

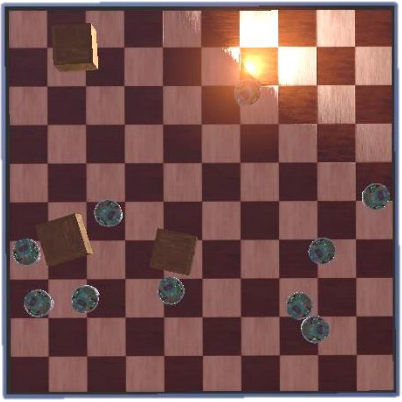

Fig. 4. A world with 10 robots, 3 boxes and a light source

"aggression" [3] and simulates how the members of the mobbing species would behave when mobbing a predator.

To implement this behavior, we programmed the controllers such that if the source of light was detected on one side, the controller would assign a higher speed to the opposite wheel causing the robot to make a sharp turn to the direction of the light source. In addition, a robot would only switch to mobbing if it received a mobbing call from another robot, or another robot acknowledged receiving its mobbing call. This Braitenbergian behavior may be interpreted as "knowledge" or "logic" [3].

*3) Worlds*

We designed simulation worlds by implementing a 1x1 meters floor. The floor is limited by four solid walls on each side. Each world contains one light source, three boxes and either ten robots (See Fig 4). We altered the translation and rotations of each object to make a variety of 10 worlds for the experiment. We used an online tool (random.org) to generate sets of random numbers for the Cartesian coordinates and rotation angle of each object (See Appendix D, See email appendix for world files).

*4) Procedure*

Each simulation would begin with the robots reading sensory inputs and detecting sources of lights and obstacles. The robots would be initialized with a default avoiding predator behavior. If a robot detected a light intensity higher than a certain threshold (marked in the codes as l_th), it would start sending a mobbing call (a string message, "must mob").

Meanwhile, if another robot received this mobbing call, they would send another call (another string message, "ok") to acknowledge that they are ready to mob the predator and the sender of the mobbing call is not alone. They would then switch their behavior from "avoiding predator" to "mobbing predator". If a robot that had sent a mobbing call previously received the acknowledgement, they would also stop switch their behavior from avoiding the predator to mobbing it.

If a robot did not receive a mobbing call, or an acknowledgement to its previous mobbing call, it would keep going with its avoiding predator behavior. While it was possible for a robot to receive a call and switch its behavior from avoiding to mobbing, nothing could make the robots switch its behavior the other way around. Finally, the robots would be avoiding obstacles at all times.

*B. Experiment Design*

We designed and conducted a within-subjects experiment to explore the effects of manipulating two experimental factors on success and quality of mobbing: i) mobbing range with three levels (no range, medium-range, low-range); and ii) group size with two levels (three robots vs ten robots).

We manipulated the mobbing range by altering the emitter range in the controller. We assigned infinite range (defined by a value of '-1' in the code) for the control condition, 0.5 meter for mid-range (covering 50% of the length of each world), and 0.1 meter for low-range (covering 10% of the length of each world).

Further, we removed six robots from each world (robot#4 to robot#10) to adjust them for three-robot conditions. We then conducted the experiment by running all 6 conditions of the experiment on each of the 10 worlds. This yielded 60 observations of the simulation such that in each observation all the robots had the same controller code.

*C. Data Collection and Analysis*

We imposed a one-minute time limit to each observation of the simulation and recorded the mobbing status (unanimous mobbing, partial mobbing, failed to mob), and the percentage of robots that participated in the mobbing. This time limit was imposed for the consistency of measurements, because in some cases the coordination was very slow.

To increase accuracy of our recordings, we programmed each robot to print a message when they switched their behavior to mobbing the predator. The robots would also print the amount of time it took them to make a mobbing decision. After one-minute we would pause the simulation and search the logs for the number of mobbing decisions.

We conducted two 2x3 repeated-measures ANOVAs to investigate the effects of group size (ten robots vs three robots) and the range of mobbing call (infinite range, mid-range, low-range) on number of unanimous mobbing attempts, and on the percentage of robots that participated in mobbing. We planned contrasts to break-down the effects in both tests. The comparisons were corrected using Bonferroni adjustment. We report descriptive analysis and findings from the ANOVA tests in the following section (See SPSS data and analysis in APPENDIX B and APPENDIX C).

V. RESULTS

We analyzed the results for 60 observations (6 experimental conditions x 10 simulation worlds). Below we detail our analyses and discuss them with respect to our research questions and hypotheses.

On average the robots failed to coordinate a mobbing in 10% of the simulations (n=6). Of the remaining 90% (n=54), half (n=27) resulted in unanimous attempts with all the robots participating in mobbing and the other half coordinated a partial mobbing with some but not all of the robots participating.

All the cases with infinite range coordinated unanimous mobbing. Unanimous mobbing rate dropped to 35% in cases with mid-range mobbing call and 0% in cases with low-range mobbing call. Half of the cases with ten robots, and 40% of the cases with three robots succeeded to coordinate a unanimous mobbing (See Figure 5).

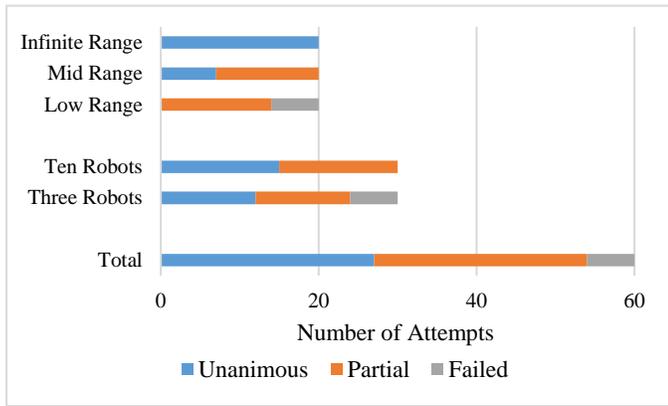

Fig. 5. Break-down of mobbing success based on experiment factors.

In an average simulation, 73.00% of the robots participated in mobbing (SD=32.62). In cases where the mobbing attempts were not unanimous, an average of 50.91% of the group participated in mobbing (SD=29.05). Descriptive statistics suggested higher participation percentage for ten-robot groups (M=82.67, SD=22.12), than three-robot groups (M=63.33, SD=38.51).

That is, the simulation results included some successful unanimous mobbing attempts, confirming our (H1a) regarding the possibility of simulating mobbing with Braitenbergian robots. Furthermore, the robots could not coordinate a unanimous mobbing in some of the scenarios or failed to mob entirely in some others. This confirms our (H1b) suggesting that it is possible to adjust the level of difficulty of mobbing in such a simulation.

Mauchly's test indicated that the assumption of sphericity had not been violated for main effect of the range of the mobbing call, $\chi^2(2) = .06$, $p = .9$, nor the interaction effect of range of the mobbing call and group size, $\chi^2(2) = 1.67$, $p = .43$ on the mobbing participation rate. This test also indicated perfect sphericity for both above mentioned effects on the number of unanimous mobbing attempts.

Our first repeated-measures ANOVA investigated the effects of experimental factors on robots' percentage of participation in mobbing. There was a significant main effect of the range of the mobbing call on the percentage of participation in mobbing, $F(2, 18) = 84.21$, $p<.001$. Contrasts revealed that robot participation was significantly lower when a range limit was imposed to the mobbing call compared to the control condition $F(1, 9)= 97.69$, $p<.001$, and it was also significantly lower for the low range compared to the mid range, $F(1,9)=71.12$, $p<.001$. This partially confirms our (H2a) and (H2b) about the effects of mobbing range on the overall success of mobbing, as it confirms success in terms of higher participation.

The first ANOVA also suggested a significant main effect of group size on the percentage of participation in mobbing, $F(1,9)=22.93$, $p<.001$, with 10-robot condition resulting in higher percentage than the 3-robot condition. This rejects our (H3b) hypothesis as we had hypothesized that robots in bigger groups will have less participation rates in mobbing.

Finally, there was a significant interaction effect between the range of the mobbing call and group size $F(2,18)=7.34$, $p<.005$. This indicates that the call range had different effects on the percentage of participation in mobbing depending on how many robots were being used. To break down this interaction, contrasts were performed comparing the range of mobbing calls to their baseline (no range imposed) and the 10-robot to the 3-robot conditions. These revealed significant interactions when imposing a call range for 3 robots compared to 10 robots $F(1,9)=22.93$, $P<00.1$.

Looking at the interaction graph (See Fig 6), this effect reflects that having an imposed call range (compared to no range) lowered the mobbing participation significantly more for 3 robots than it did for 10 robots. This confirms our (H2c) hypothesis that imposing a mobbing range will have affect mobbing success of smaller robot groups more than bigger groups. The remaining contrast revealed no significant interaction term when comparing the call range of 0.1 to 0.5 for 3 robots compared to 10 robots.

The second repeated-measures ANOVA found a significant main effect of the mobbing range on the unanimous mobbing rate $F(2,18)=67.829$, $P<.001$. Contrasts revealed that unanimous mobbing rate was significantly lower when a range limit was imposed to the mobbing call compared to the control condition $F(1, 9)= 234.05$, $p < .001$ and it was also significantly lower for the low range compared to the mid range, $F(1,9)=10.76$, $p < .01$. Once again, this partially confirms our (H2a) and (H2b) about impacts of mobbing range as it suggests impacts on unanimous mobbing attempts. Together with the results of the first ANOVA, (H2a) and (H2b) are confirmed in terms of *overall* success.

We did not find any significant effect of the number of robots on unanimous mobbing attempts; thus it does not confirm or reject our (H3a) about the impact of group size on the number of mobbing attempts. We also did not find any significant interaction effect between the mobbing call and the number of robots.

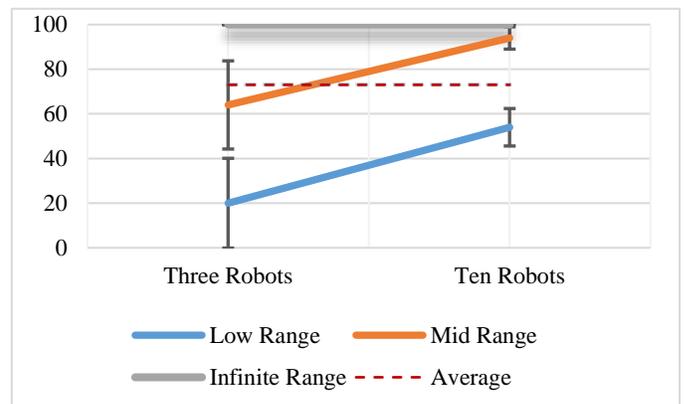

Fig. 6. Interaction effect between the the mobbing range and group siz. Error bars represent 95% confidence intervals

## VI. DISCUSSION

In this work, we simulated a dyadic anti-predator behavior with Braitenbergian robots such that the robots would avoid an inanimate predator (represented by a point-light), while trying to coordinate with other robots around them to join forces and attack the predator. If they succeeded to coordinate, they would all switch their behavior and mob the predator. We explored the effects of robots' mobbing range and group size on success of their mobbing attempts. Our results suggest significant main affects of both factors, and a significant interaction affect.

In this work, we defined "success of mobbing" in terms of both the higher numbers of successful attempts and higher percentage of group members' participation. However, our measurements do not necessarily reflect success of mobbing in real world. For example, if 9 birds (in a group of 10) mob a predator, chance are that they will be more successful in eliminating it than if if 3 birds in a group of 3 attempt to mob the same predator. In fact, there are many factors such as predator evasion, evolutionary advantages of prey or predator, etc., that could define the success or failure of mobbing attempt in real world, exploring which may be beyond the scope of this project.

As an alternative measurement, we were planning to measure the duration of mobbing (i.e. the time it took each robot to decide to mob the predator). However, while recording data we realized that we get different values for this measurement in different iterations of the same simulation. This could be rooted in different system specifications or could be a bug in the simulation software, etc. Due to this inconsistency, we decided against relying on time data in our analysis, and instead recorded participation rates which were seemingly more stable. However, our recordings of the durations are still available in the raw data (See Appendix D).

Our measurements are also limited by our 60-seconds simulation time frame. For example, we noticed that in some of the cases, more robots would join the mobbing after the first minute of simulation which is an indication of a slow coordination process, rather than a less successful mobbing attempt. This was overlooked in our data because of the arbitrary time frame that we imposed. That said, we argue that it is not practical to not have any timeframe for recording data, because as long as there are more robots wandering around that could join the mobbing, the system is not steady and the participation rates could change at any time.

We also noted that in two cases (out of 60), only one robot attempted to mob the predator. This was surpizing as it is contradictory to the nature of mobbing a predator. A potential justification to this observation is that the robot that originally sent the mobbing call missed the acknowledgement from another robot, so it refrained from participating in mobbing. That is, the robot receiving the mobbing call sent an acknowledgement and immediately proceeded to mob, but it was not received by the original sender in our 60-seconds timeframe. This is possible the mobbing range was limited in both scenarios, and the robots continuously wandered around.

Furthermore, our claims about interpretations of "fear", "aggression", "knowledge" and "logic" are merely our thought experiments and are not validated by external data. Despite the initial proposals we did not qualitatively analyze the Braitenbergian interpretations of our robot behaviors. Future work could be geared towards investigating observers' interpretations of the dyadic behavior.

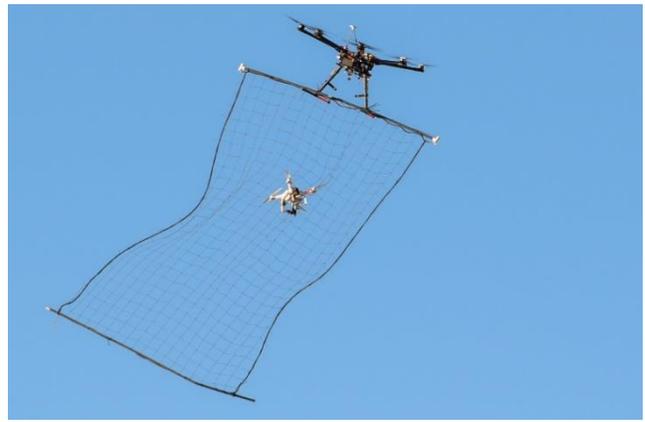

Fig. 7. United States Air Force drone hunting, source: [26]

Our experiment design was also prone to some limitations. For instance, we resorted to using external tools for randomizing the initial locations of all objects in the simulations. An improved alternative would be to randomize the initial locations using the simulation software, which we did not pursue due to the limited time we had for this project.

Finally, many of our design choices such as range values, group sizes, and number of obstacles were arbitrary, which makes the generalizability of our results very limited. Future work could explore alternative design choices or manipulate other simulation factors that went unnoticed in our study, such as aperture of the mobbing call.

We synthesized the designed behaviors using simple electronic components and processes with very little centralization. That is, robots used the same code, but were not supervised by a central node. Our work has implications for robot design and control architectures, especially in the field of autonomous defense. For example, using drones to hunt and capture other drone is a common practice (e.g. [24], see Fig 6) and this work can contribute to interdiction of drone hunting and other autonomous agent defense applications. Furthermore, bio-inspired cooperative anti-predatory behavior simulations such as this project can enable researchers to compare the process with the real animal behavior and draw insights about its characteristics. [25], [26]

## VII. CONCLUSION

In this work we simulated an evolutionary anti-predator process called mobbing in which members of a prey species join forces to cooperatively attack a predator. We built on Braitenberg's thought experiments to synthesize a dyadic avoiding/mobbing predator behavior using simple simulated robots. Our investigations suggest that factors such as robots' mobbing range and their group size can significantly impact the success of the mobbing attempts. Our work has implications for artificial life simulations of anti-predatory behaviors. It can also contribute to design and control architectures of autonomous robots, especially for defense applications.